%% file: main.tex
\documentclass[10pt,twocolumn,letterpaper]{article}
\usepackage[pagenumbers]{cvpr}
\input{preamble}
\usepackage{balance}

\definecolor{cvprblue}{rgb}{0.21,0.49,0.74}
\usepackage[breaklinks,colorlinks,allcolors=cvprblue]{hyperref}
\hypersetup{
  pdftitle={VLN-Flash: Parallel Action Decoding with Adaptive Execution for Vision-and-Language Navigation},
  pdfauthor={Ruizhe Xie, Honghua Chen, Binglin Wang, Tong Guo, Xu Tang, Puti Yan, Shaoxiong Cai}
}

\title{VLN-Flash: Parallel Action Decoding with Adaptive Execution\\ for Vision-and-Language Navigation}
\author{%
  Ruizhe Xie\textsuperscript{1,*}\quad Honghua Chen\textsuperscript{2,*}\quad
  Binglin Wang\textsuperscript{3}\quad Tong Guo\textsuperscript{4}\\
  Xu Tang\textsuperscript{5}\quad Puti Yan\textsuperscript{6,\ensuremath{\dagger}}\quad
  Shaoxiong Cai\textsuperscript{6,\ensuremath{\dagger}}\\[4pt]
  {\small\textsuperscript{1}Beijing University of Posts and Telecommunications\quad
  \textsuperscript{2}University of Electronic Science and Technology of China}\\
  {\small\textsuperscript{3}National University of Defense Technology\quad
  \textsuperscript{4}Nanyang Technological University}\\
  {\small\textsuperscript{5}Harbin Institute of Technology\quad
  \textsuperscript{6}Beihang University}\\[3pt]
  {\small\textsuperscript{*}Equal contribution.\quad
  \textsuperscript{\ensuremath{\dagger}}Corresponding authors.}
}
\date{}

\begin{document}
\maketitle

\input{sec/0_abstract}
\input{sec/1_intro}
\input{sec/2_related}
\input{fig/teaser}
\input{sec/3_method}
\input{sec/4_experiments}

\input{sec/5_conclusion}

{
    \small
    \bibliographystyle{ieeenat_fullname}
    \bibliography{main,decoder_refs}
}

\section*{Appendix}
\input{sec/appendix}

\end{document}

%% file: preamble.tex
\usepackage{algorithm}
\usepackage{algpseudocode}
\providecommand{\Require}{\State\algorithmicrequire~}

\usepackage{float}
\usepackage{amsthm}
\newtheorem{proposition}{Proposition}
\usepackage{array}
\usepackage{tabularx}
\usepackage{colortbl}

%% file: sec/0_abstract.tex
\begin{abstract}
When a vision-and-language navigation policy generates an action sequence autoregressively, the robot waits for a chain of dependent predictions. A small discrete action vocabulary makes it possible to score all short-horizon plans in parallel. We introduce VLN-Flash, a training-free decoder that evaluates the complete action-prefix tree in a single forward pass. Branch-specific attention and matched token positions preserve the original policy's conditional action distribution for each prefix in exact arithmetic. We use the score margin between the stepwise-greedy prefix and its strongest alternative at each depth to decide how many actions to execute. The robot then executes the plan with the highest joint score at the chosen depth before observing again. On matched inputs, VLN-Flash achieves a nearly $3\times$ speedup in action decoding over native autoregression on Uni-NaVid and StreamVLN. Compared with the native autoregressive decoders, it improves success rate and success weighted by path length (SPL) on R2R-CE and RxR-CE, with success gains up to 2.7 percentage points. On a Unitree GO2, VLN-Flash improves navigation success and cuts the mean wait between actions by about 160\,ms after subtracting network transmission time.
\end{abstract}

%% file: sec/1_intro.tex
\section{Introduction}
\label{sec:intro}

Vision-and-language navigation (VLN) requires an agent to follow an instruction through an unseen environment using first-person observations. Recent policies fine-tune video-based large language models to emit low-level actions as tokens~\cite{zhang2025uninavid,wei2025streamvln}, improving performance on R2R-CE and RxR-CE~\cite{anderson2018vlr,ku2020rxr,krantz2020vlrce}. In autoregressive decoding, each predicted action becomes part of the input for the next prediction. The robot waits for these successive model calls before it can execute the plan. That delay recurs each time the robot uses a new observation to plan its next actions.

Uni-NaVid and StreamVLN predict up to four actions per decision. Uni-NaVid executes at most two~\cite{zhang2025uninavid}, while StreamVLN executes the full prediction~\cite{wei2025streamvln}. In both policies, the execution limit is set in advance, regardless of how strongly the policy favors a particular plan given the current observations. This motivates using the policy's predictions to decide how much of a plan to execute before observing again.

The small navigation action space lets us enumerate all prefixes needed to score short plans. We introduce VLN-Flash (Figure~\ref{fig:overview}), a training-free decoder that packs these prefixes into a tree and scores them together in one forward pass through the navigation policy. Tree attention lets branches share the observed context while keeping each candidate's own action history. Matching the visible history and position indices to autoregressive decoding preserves the corresponding conditional distributions in exact arithmetic. The original language-model head reads out the action distribution for every branch.

Scoring the whole tree lets the robot choose how many actions to execute at each decision. Execution length determines when new observations can influence the policy. Longer blocks reduce the number of decisions along a route, while shorter blocks allow earlier replanning with fresh visual feedback. Since the tree already scores all depths, selecting a shorter execution prefix requires no additional prediction pass.

VLN-Flash uses these readouts for joint-probability plan selection and margin-based adaptive execution. Joint probabilities rank candidate action sequences using the policy's conditional predictions. To determine the execution length, we scan the score margin between the fixed stepwise-greedy branch and its strongest alternative from shallow to deep. The scan stops when the greedy branch's lead falls below a threshold, or when that branch ends. At the selected depth, the agent executes the highest-scoring branch and then observes again. The same tree scores thus determine both which plan to execute and how long to follow it.

We measure decoding on identical inputs and ablate plan selection and adaptive execution. On Uni-NaVid and StreamVLN, VLN-Flash accelerates action decoding by nearly $3\times$ over native autoregressive decoding and improves success rate (SR) and success weighted by path length (SPL) on both R2R-CE and RxR-CE, with SR gains up to 2.7 percentage points. On a Unitree GO2, the same StreamVLN policy completes 52 of 60 instructions with VLN-Flash versus 49 with autoregressive decoding.

\begin{samepage}
Our main contributions are as follows.
\begin{enumerate}
\item \textbf{One-pass action-tree decoding.} VLN-Flash scores all short-horizon action prefixes in a single forward pass without retraining, preserving autoregressive conditional distributions in exact arithmetic.
\item \textbf{Shared scores for planning and execution.} Joint probabilities rank candidate plans, and score margins determine how many actions to execute before observing again.
\item \textbf{Faster decoding and better navigation.} VLN-Flash decodes actions nearly $3\times$ faster, improves SR and SPL on both benchmarks, and reduces action waiting time on a Unitree GO2.
\end{enumerate}
\end{samepage}

%% file: sec/2_related.tex
\section{Related Work}
\label{sec:related}

\noindent\textbf{Efficient vision-and-language navigation.} Early VLN benchmarks let agents select reachable viewpoints on predefined navigation graphs~\cite{anderson2018vlr,ku2020rxr}. VLN-CE replaces graph transitions with low-level actions in continuous 3D environments~\cite{krantz2020vlrce}. In this setting, NaVid predicts the next action from egocentric video~\cite{zhang2024navid}, while Uni-NaVid predicts a short action sequence~\cite{zhang2025uninavid}. Other policies use different action representations. NaVILA issues mid-level language commands for a locomotion controller~\cite{cheng2025navila}, and OmniNav predicts continuous waypoints~\cite{xue2026omninav}. JanusVLN stores visual semantics and spatial geometry in separate compact memories~\cite{zeng2026janusvln}. StreamVLN combines SlowFast context modeling with cross-decision KV reuse~\cite{wei2025streamvln}, and Efficient-VLN processes new frames incrementally using cache reuse and contiguous rotary positions~\cite{zheng2025efficientvln}. MemVLN combines pyramidal visual memory with a learned vocabulary of single-token mid-level actions~\cite{liu2026memvln}. LiveVLN reduces robot waiting by overlapping observation processing with action execution~\cite{wang2026livevln}. We focus on speeding up plan generation once the observation history is available.

\noindent\textbf{Parallel decoding for language and actions.} Speculative decoding verifies proposals from a smaller model~\cite{leviathan2023fast}. Draft \& Verify drafts by skipping layers~\cite{zhang2024draftverify}, while Prompt Lookup retrieves candidate continuations from the input~\cite{saxena2023prompt}. Jacobi decoding updates token guesses by fixed-point iteration~\cite{santilli2023parallel}; Lookahead uses these updates to build and verify candidate continuations~\cite{fu2024lookahead}. SpecInfer and Medusa verify branches with tree attention~\cite{miao2024specinfer,cai2024medusa}, and EAGLE-2 uses draft confidence to build the tree~\cite{li2024eagle2}.

Parallel decoding has also been studied for robot actions. PD-VLA reformulates autoregressive action decoding as a nonlinear system solved by parallel fixed-point iterations~\cite{song2025pdvla}. Spec-VLA adapts draft-and-verify decoding to action tokens with a relaxed acceptance criterion~\cite{wang2025specvla}. OFT combines parallel decoding, action chunking, and continuous-action regression in a fine-tuning recipe~\cite{kim2025oft}, while Discrete Diffusion VLA generates discrete action chunks through iterative refinement~\cite{liang2026ddvla}. For short navigation plans, the small action set lets us enumerate valid prefixes directly and score the complete tree in one pass. We use the resulting joint scores to select a plan and decide when to observe again.

\noindent\textbf{Adaptive action execution.} Adaptive chunk execution determines how many predicted actions to apply before incorporating new observations. AAC estimates action entropy from sampled chunks to select the execution length at inference time~\cite{liang2026aac}. Attention-based methods derive execution horizons from the policy's internal structure. AutoHorizon uses action self-attention~\cite{wang2026autohorizon}, while cross-attention dynamics provide a truncation signal when action predictions lose grounding in the current observation~\cite{xu2026stop}. Other approaches use the predicted motion or its consequences. PACE identifies low-speed transitions in the predicted chunk as replanning boundaries~\cite{nie2026pace}; CA$^3$C uses a world model to compare the imagined consequences of candidate chunks and commits while those futures agree~\cite{li2026ca3c}. In reinforcement learning, ACH learns action values for multiple chunk lengths and selects the horizon according to the current state~\cite{shin2026ach}. We derive execution depth from the joint-score margin between navigation branches, reusing the scores produced by parallel action decoding.

%% file: fig/teaser.tex
\begin{figure*}[t]
  \centering
  \includegraphics[width=\textwidth]{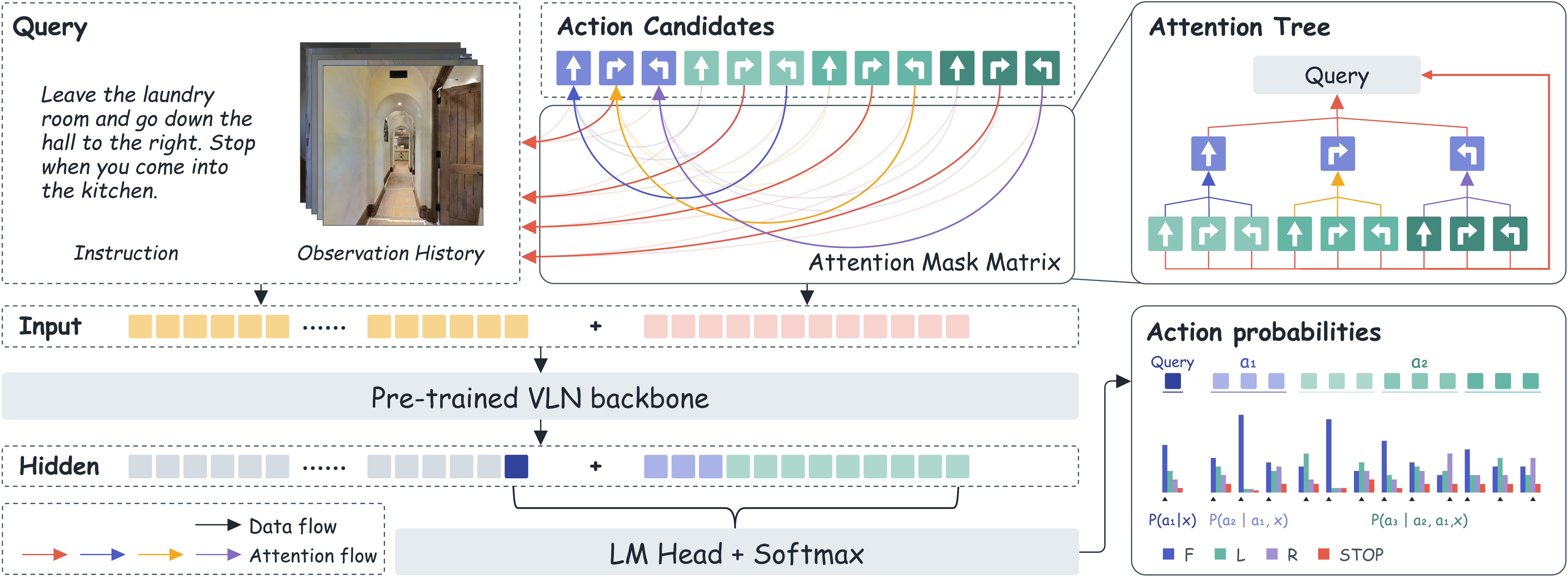}
  \caption{Overview of parallel action decoding. The instruction and observation history are formatted as the model query $x$, followed by candidate action tokens. An attention mask restricts each candidate action token to attending only to the query, its branch ancestors, and itself. The VLN backbone processes the resulting sequence in a single forward pass. The hidden state of the last query token and those of all candidate tokens are then decoded into action distributions through the shared LM head. The last query token predicts the first action, so $K=3$ uses two candidate levels. Scoring $K=4$ adds a third level with 27 candidate tokens.}
  \label{fig:overview}
\end{figure*}

%% file: sec/3_method.tex
\section{Method}
\label{sec:method}

Given the policy input $x$, VLN-Flash computes the conditional action distributions needed to score all valid plans of up to $K$ actions in one masked forward pass. These distributions support joint-probability plan selection (Section~\ref{sec:selection}) and adaptive execution (Section~\ref{sec:commit}). Figure~\ref{fig:overview} shows how the tree is packed, masked, and read out.

\subsection{Preliminaries}
\label{sec:prelim}

At decision step $t$, the instruction $I$ and egocentric video history $o_{1:t}$ form a policy input $x$ of length $L$ after visual encoding and tokenization. The fixed prompt and response header are included in $x$, so its final position predicts the first action. We obtain the action distribution $\pi(a\mid\cdot)$ by applying softmax to the language-model logits for the four actions
\begin{equation}
  \mathcal{A}=\{\texttt{forward},\texttt{left},\texttt{right},\texttt{stop}\}.
  \label{eq:actions}
\end{equation}
A \emph{plan} is a predicted action sequence, with maximum length $K=4$ here. The agent executes a selected prefix before acquiring a new observation and replanning. The native autoregressive decoder needs $h+1$ forward passes to generate $h$ action tokens and the output terminator. The first pass processes $x$ and predicts the first action; the remaining $h$ passes produce the other $h-1$ actions and the terminator.

All candidate plans at a decision condition on the same observed history. Their action prefixes describe possible continuations of that history. Once a branch is selected, the agent executes its actions and supplies the newly acquired observation to the next decision. Thus the planning horizon bounds what the decoder scores, while the execution rule determines when the policy observes again.

\begin{figure}[!ht]
  \centering
  \includegraphics[width=\linewidth]{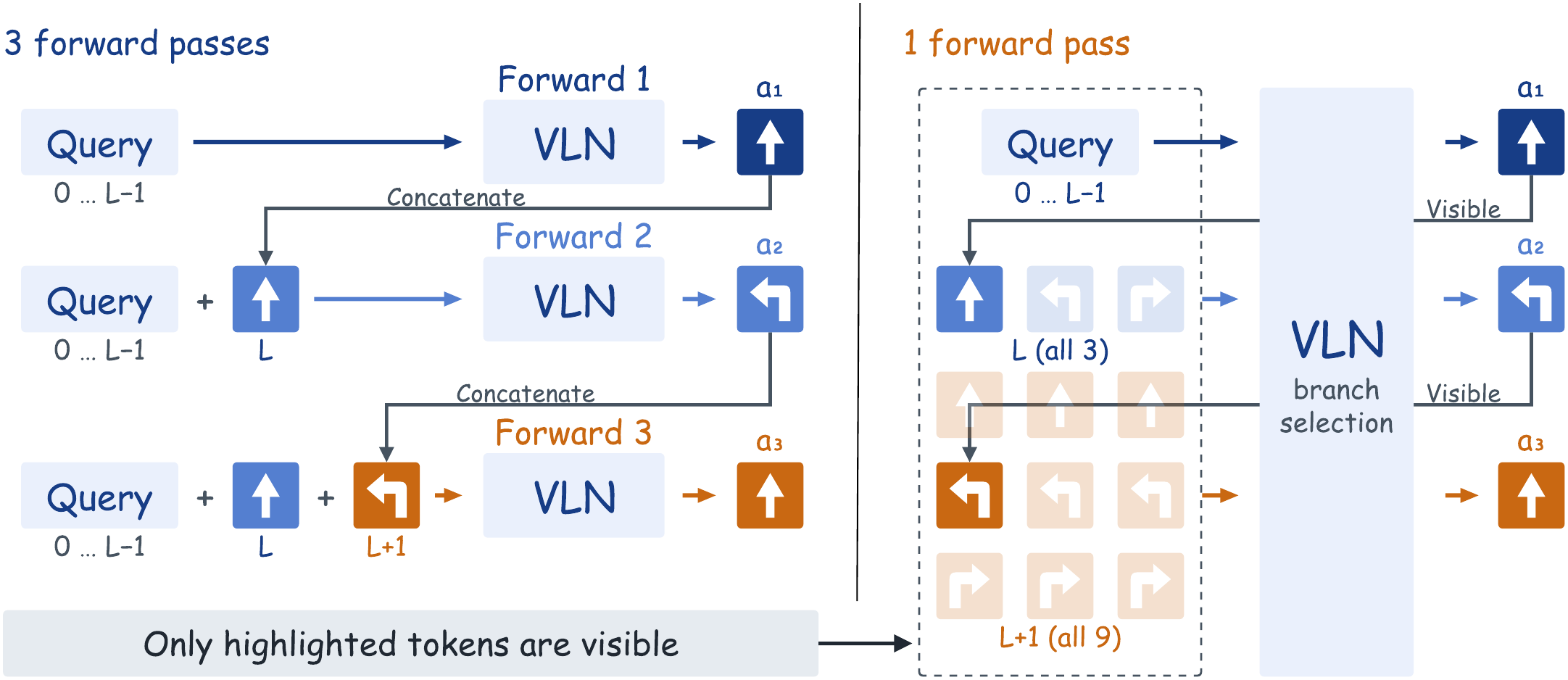}
  \caption{Comparison between autoregressive and parallel decoding. Autoregressive decoding (left) requires three sequential forward passes to predict $a_1$, $a_2$, and $a_3$, whereas parallel decoding (right) computes all candidate action distributions in a single forward pass. For each action prediction, the visible tokens (highlighted) are identical along the corresponding branch in both methods, yielding equivalent conditional action distributions in exact arithmetic.}
  \label{fig:vsauto}
\end{figure}

Figure~\ref{fig:vsauto} contrasts serial action prediction with the parallel readouts used by VLN-Flash.

\subsection{Parallel action decoding}
\label{sec:parallel}

Autoregressive decoding follows one action prefix at a time. Given a candidate prefix $s$, a causal forward on $(x,s)$ gives the distribution of the next action at its last position. VLN-Flash constructs the candidate prefixes in advance and packs them into one sequence. A tree mask then gives each readout the same context as its separate causal forward (Figure~\ref{fig:overview}).

\begin{samepage}
The last token of $x$ is the root and predicts the first action. A node at depth $k$ then predicts action $k+1$ conditioned on its prefix, so we only need to pack prefixes through depth $K-1$. Only \texttt{forward}, \texttt{left}, and \texttt{right} can extend a prefix; \texttt{stop} ends the plan. This gives $3^k$ candidate positions at depth $k$, for a total of
\begin{equation}
  |\mathcal{S}|=\sum_{k=1}^{K-1}3^k,
  \label{eq:ncand}
\end{equation}
which is 39 for $K=4$.

\end{samepage}

Figure~\ref{fig:overview} shows the $K=3$ case. The root predicts the first action, the three depth-one nodes predict the second, and the nine depth-two nodes predict the third. With $K=4$, we add 27 nodes at depth three to predict the fourth action. Each readout includes the probability of \texttt{stop}, which can end a plan at any depth.

Each nonempty prefix $s=(a_1,\dots,a_k)$ has one node $v_s$ containing its last action token $a_k$. Earlier actions are represented by its ancestors. In the packed sequence, $v_s$ attends to $x$, its ancestors $\{v_{a_{1:j}}\}_{j=1}^{k-1}$, and itself. The mask excludes all other candidate nodes and keeps attention within $x$ causal. The position index of $v_s$ is $L+k-1$, the position of $a_k$ in $(x,s)$ with zero-based indexing.

\begin{samepage}
The branch highlighted in Figure~\ref{fig:vsauto} starts with $a_1=\texttt{forward}$ and $a_2=\texttt{left}$. The node $v_{(a_1)}$ at position $L$ yields the second-action distribution $\pi(\cdot\mid x,a_1)$. Its child $v_{(a_1,a_2)}$ at position $L+1$ attends to $x$, its parent, and itself, yielding the third-action distribution $\pi(\cdot\mid x,a_1,a_2)$. A single forward pass produces both.

\end{samepage}

\begin{proposition}[Decoding equivalence]
\label{prop:equiv}
In exact arithmetic, with the specified mask and position indices, the readout state for each packed prefix $s$ equals the final state at the same readout layer in a causal forward on $(x,s)$. For the empty prefix, the readout is at the end of $x$. The same four-action head and softmax yield the same conditional distribution $\pi$.
\end{proposition}

The equivalence holds layer by layer because corresponding tokens begin with the same embeddings and positions, receive attention from the same visible prefix, and undergo the same token-wise operations. Appendix~\ref{app:proof} gives the layer-by-layer proof.

The readout uses the policy's original language-model head. Let $W_{\mathcal{A}}\in\mathbb{R}^{4\times d}$ be the rows of the head for the four action tokens, $d$ the hidden dimension, and $z_{a_{1:h}}$ the hidden state at the last token of the prefix $a_1,\dots,a_h$, where $h=0$ is the end of the model input $x$. The distribution of the action is
\begin{equation}
  \pi(\cdot\mid x,a_{1:h})=\mathrm{softmax}\bigl(W_{\mathcal{A}}\,z_{a_{1:h}}\bigr)
  \label{eq:readout}
\end{equation}
for $h=0,1,\dots,K-1$. The end of $x$ gives the distribution of step 1, and the candidate token at level $k$ gives the distribution of step $k{+}1$ under the corresponding prefix. Candidate tokens and input form a plan tree rooted at the end of $x$, whose edge to each child is labeled with the appended action $a_k$. Starting at the root and following the most likely action at each step recovers greedy decoding over the same action distributions. The readouts for the other branches are available for the comparisons in Section~\ref{sec:selection}.

Let $N_K=|\mathcal{S}|$ denote the number of candidate tokens. The scoring pass processes $L+N_K$ positions, with one copy of the visual and instruction context shared across all branches.

Within a decision, the packed sequence represents the input once. Positions in $x$ retain their causal attention within $x$ and never attend to candidate tokens, so their hidden states are the same as when the query is processed alone. Each candidate position uses those shared states alongside its own ancestors. In particular, candidates at the same depth represent separate continuations and cannot attend to one another, even when they carry the same last action token.

Each branch conditions on hypothetical actions, so scoring it requires no observations from future steps. Once the agent executes the selected actions, it incorporates a fresh observation and evaluates a new tree. This is why the execution length determines when visual feedback can affect the plan.

\subsection{Selecting actions by joint probability}
\label{sec:selection}

The plan tree keeps every branch, so first actions can be compared over longer horizons. For a branch $b=(a_1,\dots,a_k)$ we define the \emph{joint log-probability}
\begin{equation}
  L(b)=\sum_{h=1}^{k}\log\pi\bigl(a_h\mid x,a_{1:h-1}\bigr),
  \label{eq:joint}
\end{equation}
the sum of its log conditionals. The readouts provide these scores for all branches, including alternatives to the stepwise-greedy branch $g$, whose actions satisfy $g_h=\arg\max_a\pi(a\mid x,g_{1:h-1})$. We compare scores at a common depth $k$, among branches in
\begin{equation}
  \mathcal{B}_k=\{b\in\mathcal{A}^k:b_i\ne\texttt{stop}\ \text{for all }i<k\}.
  \label{eq:branches}
\end{equation}
The selected branch is $\arg\max_{b\in\mathcal{B}_k}L(b)$.

A branch ending in \texttt{stop} at depth $j$ participates in $\mathcal{B}_j$ and has no descendants; it is not carried into later comparisons. Thus every comparison uses the same number of action terms, including a terminal \texttt{stop} when present. Executing a selected \texttt{stop} ends the episode. Figure~\ref{fig:detail} works through a decision.

\subsection{Adaptive execution}
\label{sec:commit}

The greedy branch provides a reference for choosing how many actions to execute. Let $H=|g|\le K$ be its length, ending at its first \texttt{stop} or at the horizon, and write its first $k$ actions as $g_{1:k}$. Its \emph{margin} at depth $k\le H$ is
\begin{equation}
  m_k=L(g_{1:k})-\max_{b\in\mathcal{B}_k,\ b\neq g_{1:k}}L(b).
  \label{eq:margin}
\end{equation}

\begin{figure}[t]
  \centering
  \includegraphics[width=\linewidth]{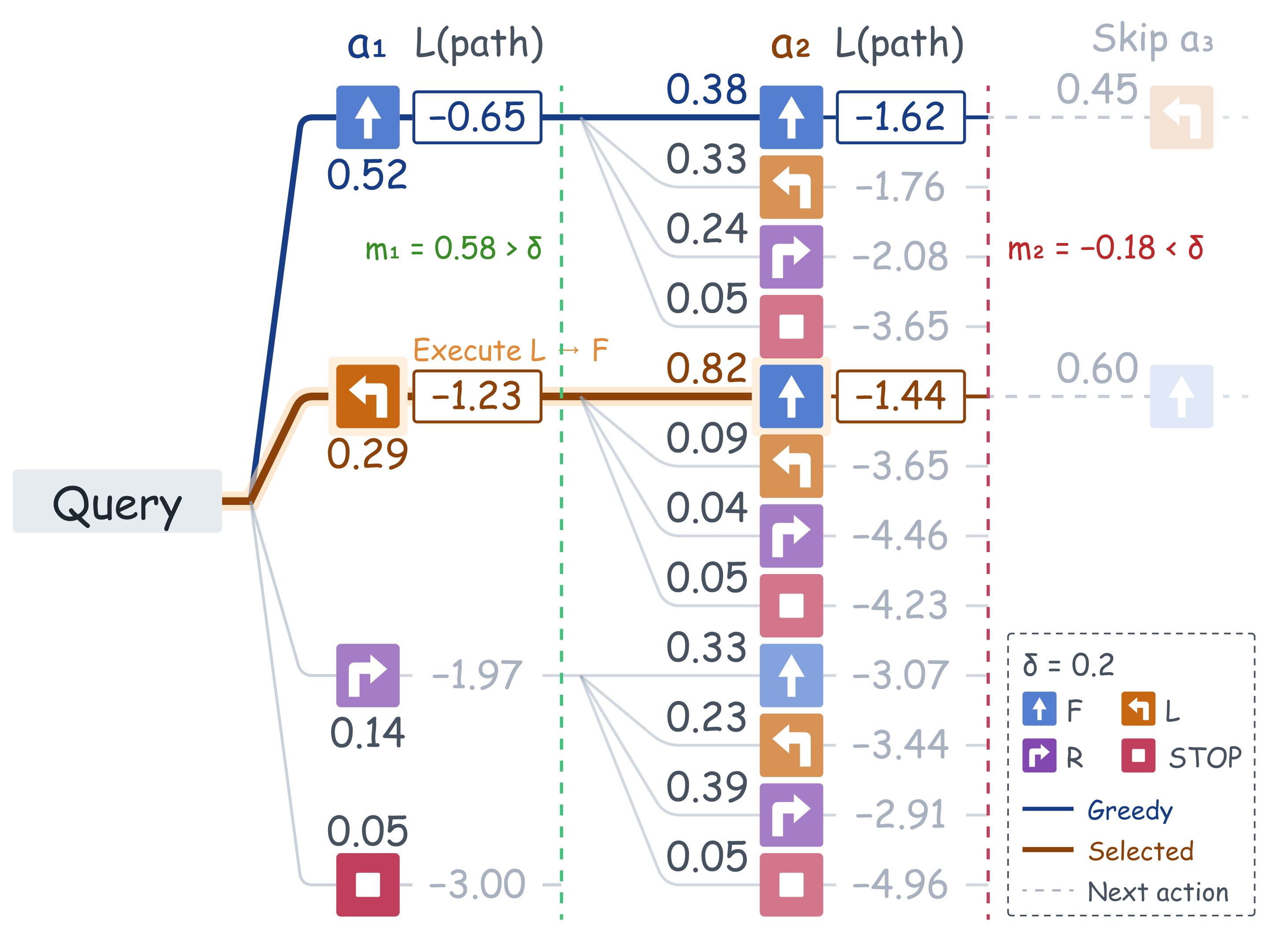}
  \caption{Adaptive execution via joint path scoring. Nodes are labeled with conditional action probabilities, while $L(\mathrm{path})$ denotes cumulative joint log-probabilities. The greedy path can be outscored by the selected path in joint probability. The scan compares the fixed greedy prefix with its strongest alternative at each depth. When their score margin falls below $\delta$, the highest-scoring branch at that depth is executed and deeper actions are skipped. Here, $m_1=0.58>\delta$ continues the scan, whereas $m_2=-0.18<\delta$ at $\delta=0.2$ selects a left turn followed by a forward step ($L\to F$).}
  \label{fig:detail}
\end{figure}

The margin shows whether the actions chosen one step at a time still form the strongest plan when scored together. A value near zero means another prefix scores similarly; a negative value means an alternative has overtaken it.

The scan starts at depth 1 and ends at the first $k$ with $m_k<\delta$, or at $H$ if no earlier depth triggers. We execute the highest-scoring branch in $\mathcal{B}_k$. It may be the greedy prefix or an alternative, and a terminal \texttt{stop} ends the episode. Algorithm~\ref{alg:commit} states the procedure.

Figure~\ref{fig:detail} shows how the margin stops the scan at depth two and the joint scores select left--forward at that depth.

The threshold $\delta$ sets the required lead. Larger values trigger earlier replanning, while negative values tolerate a competing branch temporarily scoring higher. All depths share the same scoring pass, so choosing the execution length adds no model calls. Section~\ref{sec:setup-exp} gives the threshold settings.

The horizon $K$ determines which prefixes are scored; the selected depth $k$ determines how many actions the robot executes before observing again.

%% file: sec/4_experiments.tex
\section{Experiments}
\label{sec:experiments}

We evaluate on the R2R-CE and RxR-CE val-unseen splits under the standard VLN-CE protocol. We apply VLN-Flash to two video-based large language model policies, Uni-NaVid~\cite{zhang2025uninavid} and StreamVLN~\cite{wei2025streamvln}, without retraining either model.

We first benchmark VLN-Flash against native autoregression and five training-free decoding methods on matched inputs, measuring the latency of a complete reply. We then evaluate navigation performance, ablate joint selection and adaptive execution, and measure deployment performance on a Unitree GO2.

\begin{algorithm}[t]
\caption{VLN-Flash decoding and execution}
\label{alg:commit}
\algrenewcommand{\alglinenumber}[1]{\footnotesize #1.}
\algrenewcommand{\algorithmicrequire}{\textbf{Inputs}}
\begin{algorithmic}[1]
\Require policy $\pi$; input $x$; horizon $K$; threshold $\delta$
\State Enumerate non-\texttt{stop} prefixes through depth $K-1$
\State Pack their last tokens after $x$, with ancestor attention and depth-matched positions
\State Obtain all action readouts in one policy forward
\State Accumulate $L(b)$ along each branch (Eq.~\ref{eq:joint})
\State $g \gets$ greedy branch, ending at \texttt{stop} or depth $K$
\State $H \gets |g|$
\For{$k = 1$ to $H$}
  \State $\mathcal{B}_k \gets$ length-$k$ branches with no earlier \texttt{stop}
  \State $A \gets L(g_{1:k})$
  \State $B \gets \max_{b\in\mathcal{B}_k,\,b\neq g_{1:k}}L(b)$
  \If{$A-B<\delta$ \textbf{or} $k=H$}
    \State Execute $\arg\max_{b\in\mathcal{B}_k}L(b)$
    \State \Return
  \EndIf
\EndFor
\end{algorithmic}
\end{algorithm}

\begin{table*}[!t]
\centering
\caption{Complete-reply latency across decoders. Speedup is relative to native autoregression on the same backbone.}
\label{tab:latency}
\small
\setlength{\tabcolsep}{10pt}
\begin{tabular}{@{}>{\columncolor{white}[0pt][\tabcolsep]}lrrr>{\columncolor{white}[\tabcolsep][0pt]}r@{}}
\toprule
& \multicolumn{2}{c}{\textbf{Uni-NaVid}} & \multicolumn{2}{c}{\textbf{StreamVLN}} \\
\cmidrule(lr){2-3}\cmidrule(l){4-5}
Decoder & Latency (ms) $\downarrow$ & Speedup $\uparrow$ & Latency (ms) $\downarrow$ & Speedup $\uparrow$ \\
\midrule
Native autoregressive~\cite{zhang2025uninavid,wei2025streamvln} & 150.8 & $1.00\times$ & 182.8 & $1.00\times$ \\
\midrule
Speculative~\cite{leviathan2023fast} & 218.2 & $0.69\times$ & 242.6 & $0.75\times$ \\
Draft \& Verify~\cite{zhang2024draftverify} & 136.6 & $1.10\times$ & 174.4 & $1.05\times$ \\
Prompt Lookup~\cite{saxena2023prompt} & 141.4 & $1.07\times$ & 145.9 & $1.25\times$ \\
Lookahead~\cite{fu2024lookahead} & 146.6 & $1.03\times$ & 126.5 & $1.45\times$ \\
Jacobi~\cite{santilli2023parallel} & 82.8 & $1.82\times$ & 90.2 & $2.03\times$ \\
\rowcolor{cvprblue!10}
\textbf{VLN-Flash} & \textbf{54.9} & \textbf{$2.75\times$} & \textbf{71.1} & \textbf{$2.57\times$} \\
\bottomrule
\end{tabular}
\end{table*}

\begin{table*}[!t]
\centering
\caption{Navigation performance on the val-unseen splits. The Published row lists results from the original papers; Reproduced evaluates the same checkpoints as Ours with their native decoders. SR, SPL, and OS are percentages; NE is in meters. Bold and underline mark the best and second-best results within each backbone and dataset.}
\label{tab:main}
\small
\setlength{\tabcolsep}{4.6pt}
\begin{tabular}{@{}l*{7}{r}@{\hspace{11pt}}*{7}{r}@{}}
\toprule
& \multicolumn{7}{c}{\textbf{Uni-NaVid}} & \multicolumn{7}{c}{\textbf{StreamVLN}} \\
\cmidrule(lr){2-8}\cmidrule(l){9-15}
& \multicolumn{4}{c}{R2R-CE} & \multicolumn{3}{c}{RxR-CE}
& \multicolumn{4}{c}{R2R-CE} & \multicolumn{3}{c}{RxR-CE} \\
\cmidrule(lr){2-5}\cmidrule(lr){6-8}\cmidrule(lr){9-12}\cmidrule(l){13-15}
Method & SR & SPL & NE & OS & SR & SPL & NE & SR & SPL & NE & OS & SR & SPL & NE \\
\midrule
Reproduced & \underline{52.3} & \underline{48.3} & \underline{4.98} & 57.7 & 55.1 & 44.0 & 6.01 & 55.5 & 49.7 & \underline{4.81} & \underline{63.6} & \underline{54.6} & \underline{45.8} & 5.68 \\
Published & 51.8 & 47.7 & \textbf{4.96} & \textbf{59.4} & \underline{56.1} & \underline{44.5} & \textbf{5.67} & \underline{56.4} & \underline{50.2} & 4.90 & \underline{63.6} & 54.4 & 45.4 & \underline{5.65} \\
Ours & \textbf{53.2} & \textbf{49.2} & \underline{4.98} & \underline{57.8} & \textbf{57.8} & \textbf{46.7} & \underline{5.71} & \textbf{57.5} & \textbf{51.0} & \textbf{4.73} & \textbf{65.7} & \textbf{56.2} & \textbf{46.6} & \textbf{5.57} \\
\bottomrule
\end{tabular}
\end{table*}

\subsection{Experimental setup}
\label{sec:setup-exp}

\noindent\textbf{Navigation baselines.} \emph{Reproduced} uses each backbone's released autoregressive decoding configuration. Both predict up to four actions per decision; Uni-NaVid executes at most two, while StreamVLN executes the full prediction. VLN-Flash and \emph{Reproduced} use the same checkpoint and evaluation episodes for each backbone. The \emph{Published} row lists the results from the original papers~\cite{zhang2025uninavid,wei2025streamvln}.

\noindent\textbf{Metrics.} Success rate (SR) is the fraction of episodes in which the agent stops within 3\,m of the goal. SPL weights each success by the shortest-path length divided by the larger of the shortest-path and traveled lengths. Oracle success (OS) is the fraction of trajectories that enter the success region at any point, and navigation error (NE) is the final distance to the goal in meters. We report SR, SPL and OS in percent, and decoding latency in milliseconds.

\noindent\textbf{Implementation details.} Each timed run uses one A800-80\,GB GPU with batch size one, after warm-up and with CUDA synchronization. For a given input, all decoders run on the same GPU. We retain each backbone's inference precision, using fp16 for Uni-NaVid and bf16 for StreamVLN.

\noindent\textbf{Latency measurement.} For Table~\ref{tab:latency}, we compare decoders on matched frames and histories within each backbone, using 48 Uni-NaVid inputs and 15 StreamVLN inputs with five timed runs per input. Timing starts with image tensors and prompt tokens on the GPU and includes visual encoding, reply generation, and StreamVLN’s state update for the next decision.

\noindent\textbf{Execution settings.} We set $K=4$ and split val-unseen into five folds by episode index. For each held-out fold, we select $\delta$ on the other four and pool the held-out results in Table~\ref{tab:main}. The thresholds are $0.35$ (R2R-CE) and $0.2$ (RxR-CE) for Uni-NaVid and $0$ for StreamVLN.

\subsection{Efficient decoding}
\label{sec:efficiency}

VLN-Flash has the lowest mean reply latency on both backbones (Table~\ref{tab:latency}). It reduces mean latency from 150.8 to 54.9\,ms on Uni-NaVid ($2.75\times$) and from 182.8 to 71.1\,ms on StreamVLN ($2.57\times$). Compared with Jacobi, the fastest alternative, VLN-Flash reduces latency by 33.8\% on Uni-NaVid and 21.2\% on StreamVLN.

Jacobi iterates on token guesses, Lookahead and Prompt Lookup verify candidates, and Draft \& Verify adds skipped-layer drafting passes. Speculative decoding uses TinyLlama~\cite{zhang2024tinyllama} and Qwen2 as draft models. On StreamVLN, it peaks at 16.39\,GiB, versus 15.41\,GiB for VLN-Flash. Appendix~\ref{app:protocol} details each decoder.

\begin{table}[!t]
\centering
\caption{Mean latency (ms) to generate a complete reply at each prediction horizon $K$.}
\label{tab:horizon-latency}
\small
\setlength{\tabcolsep}{4pt}
\begin{tabular}{@{}crrrr@{}}
\toprule
& \multicolumn{2}{c}{\textbf{Uni-NaVid}} & \multicolumn{2}{c}{\textbf{StreamVLN}} \\
\cmidrule(lr){2-3}\cmidrule(l){4-5}
$K$ & Autoregressive & VLN-Flash & Autoregressive & VLN-Flash \\
\midrule
$2$ & 107.0 & 50.1 & 149.4 & 68.4 \\
$3$ & 132.5 & 51.5 & 168.3 & 69.4 \\
$4$ & 152.5 & 55.3 & 186.4 & 71.6 \\
\bottomrule
\end{tabular}
\end{table}

\noindent\textbf{Effect of prediction horizon.} We replay the same inputs at $K=2,3,4$ and use greedy action selection in both decoders (Table~\ref{tab:horizon-latency}). The two decoders return the same number of actions for each input at each horizon. Raising $K$ from 2 to 4 adds 37--46\,ms to autoregression, while VLN-Flash adds about 3--5\,ms by scoring the extra candidate levels in the same forward pass.

\input{fig/history_stability_table}

\noindent\textbf{Effect of cached history.} We also preload the history cache and current query before timing to isolate action decoding from scene preprocessing. Across the standard 32-action window, the absolute time saved remains between 53.8 and 54.3\,ms as the cached context grows from 533 to 1627 positions (Table~\ref{tab:history-stability}). Longer context adds attention cost to both decoders, but does not erode the per-decision reduction from parallel action decoding.

\subsection{Navigation performance}
\label{sec:nav}

VLN-Flash improves both SR and SPL over the reproduced and published baselines on all four backbone--dataset pairs (Table~\ref{tab:main}). Uni-NaVid gains most on RxR-CE, with SR increasing from 55.1 to 57.8 and SPL from 44.0 to 46.7. StreamVLN gains most on R2R-CE, improving SR by 2.0 percentage points and SPL by 1.3.

\subsection{Ablation and analysis}
\label{sec:analysis}

\begin{table}[!t]
\centering
\caption{Joint selection and adaptive execution. Joint only selects the plan with the highest joint score at a fixed depth (two actions for Uni-NaVid and four for StreamVLN); Ours selects the depth adaptively. Metrics and rank markers follow Table~\ref{tab:main}.}
\label{tab:ablation}
\small
\setlength{\tabcolsep}{3.4pt}
\begin{tabular}{@{}lrrrrrrr@{}}
\toprule
& \multicolumn{4}{c}{R2R-CE} & \multicolumn{3}{c}{RxR-CE} \\
\cmidrule(lr){2-5}\cmidrule(l){6-8}
Variant & SR & SPL & NE & OS & SR & SPL & NE \\
\midrule
\multicolumn{8}{@{}l}{\textbf{Uni-NaVid}} \\
Reproduced & \underline{52.3} & 48.3 & \underline{4.98} & \underline{57.7} & 55.1 & 44.0 & 6.01 \\
Joint only & 51.8 & \underline{48.6} & \textbf{4.96} & 57.5 & \underline{56.2} & \underline{46.1} & \underline{5.92} \\
Ours & \textbf{53.2} & \textbf{49.2} & \underline{4.98} & \textbf{57.8} & \textbf{57.8} & \textbf{46.7} & \textbf{5.71} \\
\midrule
\multicolumn{8}{@{}l}{\textbf{StreamVLN}} \\
Reproduced & 55.5 & 49.7 & 4.81 & 63.6 & 54.6 & 45.8 & 5.68 \\
Joint only & \underline{56.9} & \underline{50.8} & \underline{4.79} & \underline{64.4} & \underline{55.9} & \underline{46.5} & \underline{5.65} \\
Ours & \textbf{57.5} & \textbf{51.0} & \textbf{4.73} & \textbf{65.7} & \textbf{56.2} & \textbf{46.6} & \textbf{5.57} \\
\bottomrule
\end{tabular}
\end{table}

We vary plan scoring in Table~\ref{tab:ablation} and execution length in Table~\ref{tab:fixedk}.

\noindent\textbf{Joint selection and adaptation.} For StreamVLN, joint selection raises SR from 55.5 to 56.9 on R2R-CE and from 54.6 to 55.9 on RxR-CE. Adaptive execution adds a further improvement to 57.5 and 56.2. On Uni-NaVid, joint selection changes SR from 52.3 to 51.8 on R2R-CE and from 55.1 to 56.2 on RxR-CE. Adaptive execution raises these to 53.2 and 57.8, respectively.

\begin{table}[!t]
\centering
\caption{Adaptive versus fixed execution (SR/SPL, \%). All variants select plans by joint score. Fixed variants execute up to $k$ actions and end early on \texttt{stop}. Avg.\,$k$ refers to adaptive execution and excludes the final decision ending in \texttt{stop}. Bold and underline mark the best and second-best SR/SPL scores.}
\label{tab:fixedk}
\small
\setlength{\tabcolsep}{5.6pt}
\begin{tabular}{@{}lrrrr@{}}
\toprule
& \multicolumn{2}{c}{Uni-NaVid} & \multicolumn{2}{c}{StreamVLN} \\
\cmidrule(lr){2-3}\cmidrule(lr){4-5}
Execution & R2R-CE & RxR-CE & R2R-CE & RxR-CE \\
\midrule
$k=2$ & \underline{51.8/48.6} & \underline{56.2/46.1} & 54.3/48.2 & 53.4/44.7 \\
$k=3$ & 51.6/48.4 & 55.3/44.6 & 54.5/49.1 & 54.6/45.2 \\
$k=4$ & 51.2/48.1 & 54.9/44.4 & \underline{56.9/50.8} & \underline{55.9/46.5} \\
Adaptive & \textbf{53.2/49.2} & \textbf{57.8/46.7} & \textbf{57.5/51.0} & \textbf{56.2/46.6} \\
Avg.\,$k$ & 3.1 & 3.4 & 3.8 & 3.8 \\
\bottomrule
\end{tabular}
\end{table}

\begin{figure}[!t]
  \centering
  \includegraphics[width=\linewidth]{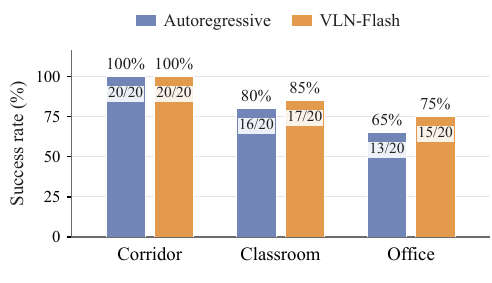}
  \caption{Success rates on the GO2 over 20 trials per environment, using the same StreamVLN policy, instructions, and start poses. Bar labels show success rates and successful-trial counts.}
  \label{fig:real-success}
\end{figure}

\noindent\textbf{Adaptive versus fixed execution.} The best fixed length is two actions for Uni-NaVid and four for StreamVLN (Table~\ref{tab:fixedk}). Adaptive execution achieves the highest SR and SPL on all four backbone--dataset pairs. On Uni-NaVid R2R-CE, it averages 3.1 actions per decision and improves SR/SPL from $51.8/48.6$ at the best fixed length to $53.2/49.2$. On StreamVLN, it averages 3.8 actions on both benchmarks while allowing earlier observations when the margin is small.

\noindent\textbf{Numerical agreement of parallel readouts.} Proposition~\ref{prop:equiv} establishes equality in exact arithmetic. We also compare tree readouts with causal forwards of the same prefixes at $K=4$ in each backbone's inference precision. Top-1 agreement is $800/800$ in each Uni-NaVid setting and $778/780$ on StreamVLN. Appendix~\ref{app:equiv} reports the logit errors and distribution distances.

Promoting StreamVLN to fp32 gives $780/780$ top-1 agreement and a maximum KL divergence of $1.72\times10^{-6}$, compared with $7.93\times10^{-4}$ in its native bf16 setting. The~two bf16 ranking differences occur at closely scored alternatives.

\begin{figure*}[!t]
  \centering
  \includegraphics[width=\textwidth]{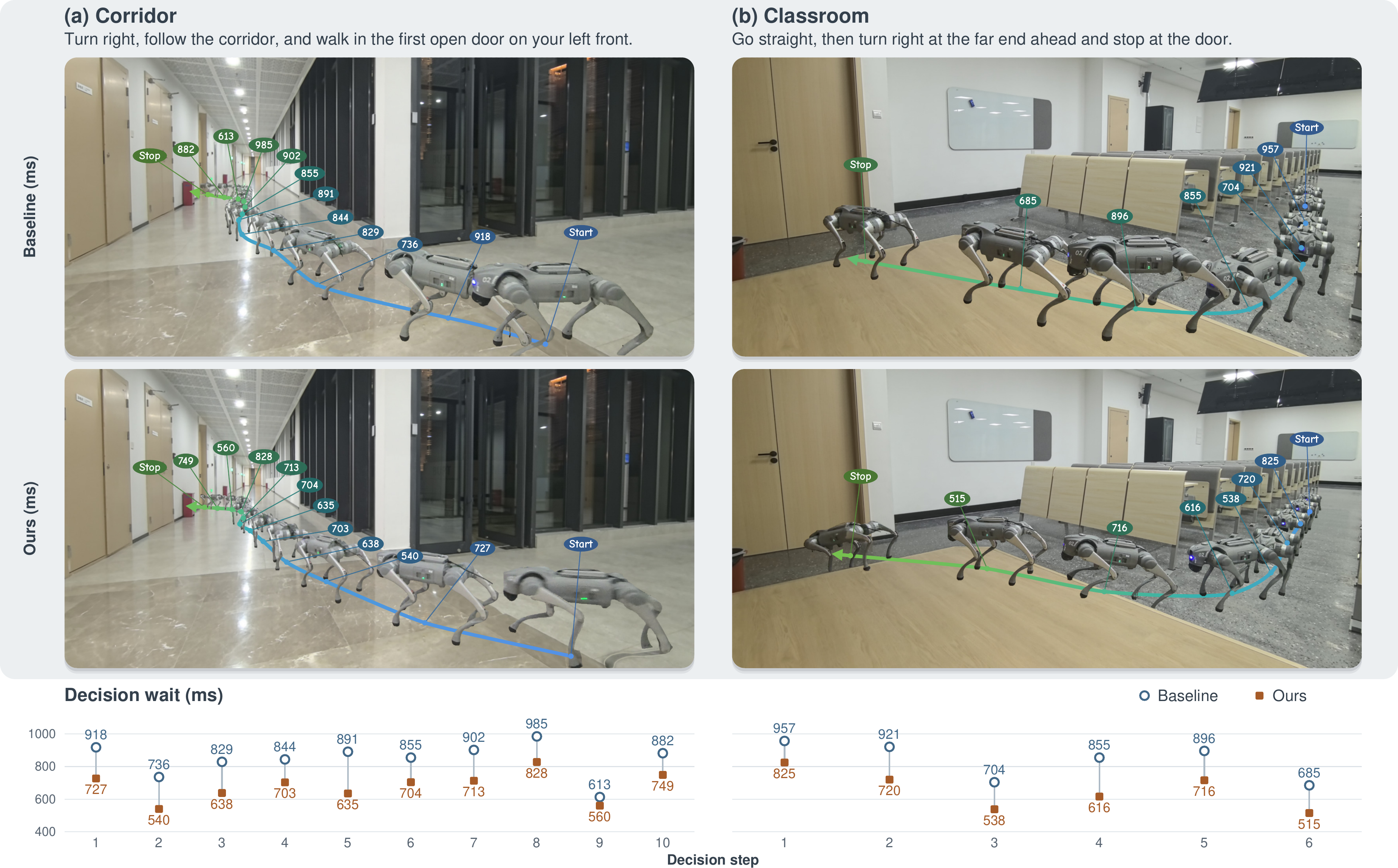}
  \caption{Real-world deployment trajectories on a Unitree GO2. The left and right scenes show a corridor and a classroom, respectively. In each scene, both methods start from the same position and receive the same instruction shown above. \textbf{Baseline} and \textbf{Ours} correspond to the autoregressive and parallel decoding variants of StreamVLN, respectively. Across the decision steps shown in these two scenes, our method reduces the average wait by 172\,ms.}
  \label{fig:real}
\end{figure*}

\begin{figure*}[!t]
  \centering
  \includegraphics[width=\textwidth]{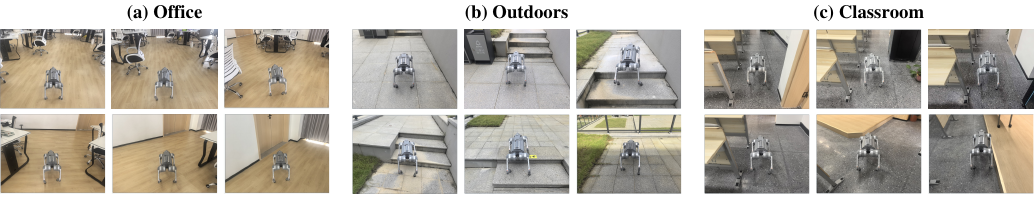}
  \caption{Additional Unitree GO2 deployments. (a) Office. ``Turn left ahead, walk to the door, and stop in front of it.'' (b) Outdoors. ``Go straight along this road, climb the stairs, and stop before the glass railing.'' (c) Classroom. ``Go straight down this road to the end, and stop beside the table at the podium on your left.''}
  \label{fig:deployment-examples}
\end{figure*}
\subsection{Real-world deployment}
\label{sec:real}

We deploy StreamVLN on a Unitree GO2 and compare its native autoregressive decoder with VLN-Flash in separate runs with matched instructions and start poses. An onboard Jetson Orin NX 16\,GB captures RGB frames and sends velocity and stop commands through ROS2. The navigation policy runs on an A800-80\,GB cloud GPU.

Following NaVILA~\cite{cheng2025navila}, we test 20 instructions each in a corridor with indoor and outdoor sections, a classroom, and an office.

\noindent\textbf{Navigation success.} A trial succeeds if the robot stops within 1.5\,m of the target. VLN-Flash completes 52 of 60 trials (86.7\%), compared with 49 (81.7\%) for native autoregression (Figure~\ref{fig:real-success}). Both methods complete all 20 corridor trials. VLN-Flash completes 17 classroom and 15 office trials, compared with 16 and 13 for native autoregression.

\noindent\textbf{Action waiting time.} Across the robot trials, VLN-Flash reduces the mean wait from 736\,ms to 572\,ms, a reduction of about 160\,ms. We measure from the end of one action to the start of the next, including frame acquisition, image processing, and policy inference, and subtract network transmission time. Figure~\ref{fig:real} illustrates decision waits from separate runs in the corridor and classroom. Figure~\ref{fig:deployment-examples} shows the robot approaching and stopping at indoor and outdoor landmarks.

%% file: fig/history_stability_table.tex
\begin{table}[t]
\centering
\caption{StreamVLN action-decoding latency with a preloaded cache. Timing starts with
the cached history and current query ready, excluding visual preprocessing and cache
construction. For each trajectory, the median of ten timed repetitions is used;
values average the three trajectories.}
\label{tab:history-stability}
\small
\setlength{\tabcolsep}{0pt}
\begin{tabularx}{\linewidth}{@{}*{4}{>{\centering\arraybackslash}X}@{}}
\toprule
\begin{tabular}[b]{@{}c@{}}Cached history\\[-1pt](positions)\end{tabular}
& \begin{tabular}[b]{@{}c@{}}Native AR\\[-1pt](ms)\end{tabular}
& \begin{tabular}[b]{@{}c@{}}VLN-Flash\\[-1pt](ms)\end{tabular}
& \begin{tabular}[b]{@{}c@{}}Saved\\[-1pt](ms)\end{tabular} \\
\midrule
533 & 85.8 & 32.0 & 53.8 \\
752 & 86.8 & 32.5 & 54.3 \\
1190 & 87.9 & 34.0 & 53.9 \\
1627 & 89.6 & 35.6 & 54.0 \\
\bottomrule
\end{tabularx}
\end{table}

%% file: sec/5_conclusion.tex
\section{Conclusion}
\label{sec:conclusion}

VLN-Flash speeds up navigation decoding by scoring the full tree of short-horizon action prefixes in one forward pass. The branches share the observation context while retaining their own action histories and the corresponding autoregressive position indices. The resulting scores also guide plan selection and determine how many actions to execute before observing again.

\begin{samepage}
This design speeds up action decoding by nearly $3\times$ over native autoregression on Uni-NaVid and StreamVLN. On a Unitree GO2, the mean wait between actions drops by about 160\,ms. With adaptive execution, SR and SPL also improve on R2R-CE and RxR-CE.

\end{samepage}

%% file: sec/appendix.tex
\appendix
\setcounter{table}{0}
\setcounter{figure}{0}
\setcounter{equation}{0}
\renewcommand{\theHtable}{appendix.\arabic{table}}
\renewcommand{\theHfigure}{appendix.\arabic{figure}}
\renewcommand{\theHequation}{appendix.\arabic{equation}}
\renewcommand{\thetable}{S\arabic{table}}
\renewcommand{\thefigure}{S\arabic{figure}}
\renewcommand{\theequation}{S\arabic{equation}}

\section{Implementation details}
\label{app:impl}

For \texttt{stop}, \texttt{forward}, \texttt{left}, and \texttt{right}, the token IDs are $(5040,6375,2175,1492)$ for Uni-NaVid and $(50669,76286,71858,51018)$ for StreamVLN. Uni-NaVid uses hidden layer 31 for the action readout.

\section{Decoding equivalence}
\label{app:proof}

\subsection{Proof of Proposition~\ref{prop:equiv}}

\begin{proof}
Fix a packed prefix $s=(a_1,\ldots,a_k)$, allowing $k=0$, and let $y=(x,s)$. Index $y$ from zero and map its positions to the packed input by
\begin{equation}
  \phi(i)=
  \begin{cases}
    i, & 0\le i<L,\\
    \operatorname{idx}(v_{a_{1:j}}), & i=L+j-1,\quad 1\le j\le k.
  \end{cases}
\end{equation}
Here $\operatorname{idx}(v)$ denotes the packed position of node $v$. The position index at $\phi(i)$ is $i$, and the mask gives it the visible set
\begin{equation}
  \mathcal V_{\mathrm{tree}}(\phi(i))
     = \{\phi(j)\mid 0\le j\le i\}.
  \label{eq:supp-visible}
\end{equation}
Thus every position in the causal input sees exactly the corresponding positions in the tree.

Write $h_i^{(\ell)}$ and $\widetilde h_{\phi(i)}^{(\ell)}$ for causal and packed hidden states after layer $\ell$. The input tokens and position indices agree, so $\widetilde h_{\phi(i)}^{(0)}=h_i^{(0)}$ for all $i$. Suppose equality holds after layer $\ell$. In each attention head of the next layer, corresponding query, key and value projections then agree, as do their positional transformations. Write $e_{ij}$ for a head's attention logit after applying position information and before masking, and $u_j$ for its value vector. By Eq.~\ref{eq:supp-visible}, the packed output $\widetilde o$ satisfies
\begin{equation}
\begin{aligned}
 \widetilde o_{\phi(i)}
 &=\sum_{j=0}^{i}
   \frac{\exp(\widetilde e_{\phi(i),\phi(j)})}
        {\sum_{r=0}^{i}\exp(\widetilde e_{\phi(i),\phi(r)})}
   \widetilde u_{\phi(j)}\\
 &=\sum_{j=0}^{i}
   \frac{\exp(e_{ij})}{\sum_{r=0}^{i}\exp(e_{ir})}\,u_j
   =o_i .
\end{aligned}
\end{equation}
The head outputs agree, and the output projection, residual updates, normalization and feed-forward operations preserve equality at each position. Hence $\widetilde h_{\phi(i)}^{(\ell+1)}=h_i^{(\ell+1)}$ for every $i$.

By induction the readout states $z_i$ and $\widetilde z_{\phi(i)}$ agree. Set $i=L+k-1$, which is $L-1$ for the first-action readout at $k=0$. The shared action head gives
\begin{equation}
  \operatorname{softmax}(W_{\mathcal A}\widetilde z_{\phi(i)})
    =\operatorname{softmax}(W_{\mathcal A}z_i).\qedhere
\end{equation}
\end{proof}

\subsection{Numerical comparison}
\label{app:equiv}

We compare the four-action readouts from the packed tree and from causal forwards of the same prefixes. Each configuration uses 20 decisions from R2R-CE at $K=4$ in evaluation mode on one A800-80\,GB GPU. For Uni-NaVid, each decision contributes the first-action readout and all 39 continuation readouts. We run it in fp16 with and without prefix reuse. For StreamVLN, we compare the 39 continuation readouts per decision with prefix reuse in bf16 and fp32.

\begin{table}[H]
\centering
\caption{Agreement between tree and causal readouts at $K=4$. $N$ is the number of action distributions, $\max|\Delta\ell|$ is the largest absolute difference across the four action logits, and Top-1 counts matching most likely actions.}
\label{tab:equiv-supp}
\footnotesize
\setlength{\tabcolsep}{3.5pt}
\begin{tabularx}{\columnwidth}{@{}Xllrrr@{}}
\toprule
Backbone & Precision & Cache & $N$ & $\max|\Delta\ell|$ & Top-1 \\
\midrule
Uni-NaVid & fp16 & Off & 800 & $4.88\times10^{-2}$ & 800/800 \\
Uni-NaVid & fp16 & On & 800 & $4.30\times10^{-2}$ & 800/800 \\
StreamVLN & bf16 & On & 780 & $2.74\times10^{-1}$ & 778/780 \\
StreamVLN & fp32 & On & 780 & $9.16\times10^{-3}$ & 780/780 \\
\bottomrule
\end{tabularx}
\end{table}

On Uni-NaVid, the mean KL divergence from the causal to the tree distributions is $5.17\times10^{-6}$ without prefix reuse and $5.41\times10^{-6}$ with reuse. The two StreamVLN bf16 mismatches in Table~\ref{tab:equiv-supp} have gaps of 0.0008 and 0.032 between the two highest causal logits. In fp32, all StreamVLN top-1 predictions agree, and the maximum KL divergence falls from $7.93\times10^{-4}$ to $1.72\times10^{-6}$.

\section{Decoding latency replay}
\label{app:protocol}
\label{app:repro}

\subsection{Replay protocol}
\label{app:complete-reply}

For the latency comparison in Table~\ref{tab:latency}, all decoders replay the same frames and observation histories within each backbone. Each input is run on the same A800-80\,GB GPU at batch size one, with two warm-up runs and five timed runs per decoder. GPU synchronization brackets each call. The timer starts with preprocessed images and prompt tokens on the GPU and includes visual encoding and complete reply generation. The full-vocabulary baselines use deterministic generation, so their replies can be compared token by token with native autoregression.

\balance
\noindent\textbf{Uni-NaVid.} We replay 48 decisions from eight R2R-CE val-unseen trajectories in fp16. Each decoder processes the full query from the same visual history. The five full-vocabulary comparison decoders match all 240 native replies per method. VLN-Flash averages 3.08 actions per reply with $\delta=0.35$.

\noindent\textbf{StreamVLN.} We replay 15 decisions from three trajectories in bf16 with SDPA. All decoders start from the same continuation state. The full-vocabulary baselines use the native repetition penalty of 1.05. VLN-Flash applies the same penalty to actions already present in each candidate prefix before normalization and selects with $K=4$ and $\delta=0$.

Speculative decoding uses TinyLlama-1.1B-Chat-v1.0 for Uni-NaVid and Qwen2-0.5B-Instruct for StreamVLN. The text draft proposes one token per round; the target verifies it with the full visual input.

\subsection{Baseline configurations}

Jacobi initializes its token guesses from the preceding reply. Lookahead uses level 4, window 5, and five guesses. Prompt Lookup matches up to two tokens and proposes at most three. Draft \& Verify drafts up to three tokens, skipping the attention and MLP blocks at layers 2--5, 7--10, 12--15, 17--20, and 22--25 (zero-based). We choose the settings with the lowest mean latency on six separate StreamVLN inputs and use those settings for Uni-NaVid.

\subsection{StreamVLN replay results}

All methods return 3.87 actions per reply on average. VLN-Flash and the four verification baselines match every native reply token for token across 75 timed runs on 15 continuation inputs. Jacobi matches 70 of 75 replies in bf16. The five differences come from repeated runs on one input, for which Jacobi and native decoding agree in fp32.

Table~\ref{tab:decoder-cost} reports the forward counts and memory use. VLN-Flash scores the tree in one forward pass, then processes the selected reply in a short pass to update the state for the next decision. Both passes are included in the latency reported in Table~\ref{tab:latency}.

\begin{table}[H]
\centering
\caption{StreamVLN replay costs. Target and Draft are mean target and draft forward counts per reply. A dash marks an uninstrumented count. Memory is peak GPU allocation in GiB.}
\label{tab:decoder-cost}
\footnotesize
\setlength{\tabcolsep}{4pt}
\begin{tabularx}{\columnwidth}{>{\raggedright\arraybackslash}Xrrr}
\toprule
Decoder & Target & Draft & GiB \\
\midrule
StreamVLN (autoregressive) & -- & -- & 15.44 \\
Speculative (0.5B draft) & 5.73 & 5.93 & 16.39 \\
Draft \& Verify & 5.00 & 6.13 & 15.52 \\
Prompt Lookup & 5.60 & 0.00 & 15.49 \\
Lookahead & 4.41 & 0.00 & 15.51 \\
Jacobi & 2.93 & 0.00 & 15.60 \\
\textbf{VLN-Flash} & 2.00 & 0.00 & 15.41 \\
\bottomrule
\end{tabularx}
\end{table}

\subsection{Prediction horizon}
\label{app:horizon}

For Table~\ref{tab:horizon-latency}, we replay the same inputs with stepwise greedy action selection in both decoders. At $K=2,3$, we cap native generation after $K$ actions and generate the reply terminator. At $K=4$ we use the native replies. VLN-Flash packs prefixes through depth $K-1$ and reads out the greedy path. Both methods return the same number of actions for each input at every horizon. The action sequences match on all inputs at $K=2,3$. At $K=4$, they match on 47 of 48 Uni-NaVid inputs and all 15 StreamVLN inputs.